\documentclass[sigconf]{acmart}
\AtBeginDocument{%
  \providecommand\BibTeX{{%
    \normalfont B\kern-0.5em{\scshape i\kern-0.25em b}\kern-0.8em\TeX}}}

\setcopyright{acmcopyright}
\copyrightyear{2018}
\acmYear{2018}
\acmDOI{XXXXXXX.XXXXXXX}

\acmConference[Conference acronym 'XX]{Make sure to enter the correct
  conference title from your rights confirmation emai}{June 03--05,
  2018}{Woodstock, NY}
\acmPrice{15.00}
\acmISBN{978-1-4503-XXXX-X/18/06}

\usepackage{amsmath, amsfonts, amssymb, amsthm, xspace, color, multirow}
\newcommand{\etc}{\emph{etc.}\xspace} 
\newcommand{\ie}{\emph{i.e.}\xspace} 
\newcommand{\eg}{\emph{e.g.\xspace}} 

\newcommand{\nop}[1]{}

\begin{document}
\captionsetup{font={small}}
\setcopyright{none}
\title{Label-Efficient Interactive Time-Series Anomaly Detection}

\author{Hong Guo}
\authornote{The authors contributed equally to this work. Last names are sorted by alphabet.}
\authornote{The work was done when the authors did internship at Microsoft.}
\affiliation{%
  \institution{Tsinghua University}
  \city{Beijing}
  \country{China}
}
\email{guoh20@mails.tsinghua.edu.cn}

\author{Yujing Wang}
\authornotemark[1]
\affiliation{%
  \institution{Microsoft}
  \city{Beijing}
  \country{China}}
\email{yujwang@microsoft.com}

\author{Jieyu Zhang}
\authornotemark[1]
\authornotemark[2]
\affiliation{%
  \institution{University of Washington}
  \country{USA}
}
\email{jieyuz2@cs.washington.edu}

\author{Zhengjie Lin}
\authornotemark[2]
\author{Yunhai Tong}
\affiliation{%
  \institution{School of Artificial Intelligence, Peking University}
  \city{Beijing}
  \country{China}
  }
\email{zhengjielin@stu.pku.edu.cn, yhtong@pku.edu.cn}

\author{Lei Yang}
\authornotemark[2]
\affiliation{%
  \institution{Peking University}
  \city{Beijing}
  \country{China}
  }
\email{yang_lei@pku.edu.cn}

\author{Luoxing Xiong}
\author{Congrui Huang}
\authornote{Corresponding author.}
\affiliation{%
  \institution{Microsoft}
  \city{Beijing}
  \country{China}}
\email{{luxiong,conhua}@microsoft.com}


\renewcommand{\shortauthors}{Guo and Wang, et al.}

\begin{abstract}

Time-series anomaly detection is an important task and has been widely applied in the industry. Since manual data annotation is expensive and inefficient, most applications adopt unsupervised anomaly detection methods, but the results are usually sub-optimal and unsatisfactory to end customers. Weak supervision is a promising paradigm for obtaining considerable labels in a low-cost way, which enables the customers to label data by writing heuristic rules rather than annotating each instance individually. However, in the time-series domain, it is hard for people to write reasonable labeling functions as the time-series data is numerically continuous and difficult to be understood.
In this paper, we propose a Label-Efficient Interactive Time-Series Anomaly Detection (LEIAD) system, which enables a user to improve the results of unsupervised anomaly detection by performing only a small amount of interactions with the system. To achieve this goal, the system integrates weak supervision and active learning collaboratively while generating labeling functions automatically using only a few labeled data. All of these techniques are complementary and can promote each other in a reinforced manner.
We conduct experiments on three time-series anomaly detection datasets, demonstrating that the proposed system is superior to existing solutions in both weak supervision and active learning areas. Also, the system has been tested in a real scenario in industry to show its practicality.
\end{abstract}





\maketitle

\section{Introduction}

Time series data is an important form of structured data and has become a research hotspot in recent years. Among numerous time series tasks, anomaly detection is an important one due to its wide existence and usefulness in various industries. Anomalies are the data points which deviate remarkably from the general distribution of the whole dataset and occupy only a small portion of the data~\cite{braei2020anomaly}. The task of anomaly detection is to develop methods to find these anomalous points. In the past, researchers have used statistical methods to establish anomaly detection methods. For example, as early as 1977, Tukey \cite{tukey1977exploratory} introduced the application of statistical methods to time series anomaly detection. Nowadays, with the popularity of artificial neural networks in computer vision and natural language processing, enthusiasm has also shifted to deep learning-based methods in the field of time series anomaly detection. For instance, Munir \cite{munir2018deepant} and Malhotra \cite{malhotra2015long}, respectively, utilized Convolutional Neural Networks (CNN) and stacked Long Short Term Memory (LSTM) networks to model and predict the behavior of a set of time series, thereby detecting deviations from normal behavior. These approaches have all yielded compelling results, leading to influential contributions.

Practically, with the surge in data volume recently, the lack of labels has become the primary issue. Machine learning methods of time series analysis mostly rely on large amounts of high-quality labeled data, which are not easily available in real scenarios. In order to ensure the reliability and accuracy of labels, people often obtain labels by manual annotation, which is tedious and time-consuming. The limited manual labeling ability cannot meet the labeling needs of the increasingly large training datasets in the industry, limiting the development and application of anomaly detection algorithms in the industry. Our efforts aim to relieve the pressure of manual data annotation and develop more precise and efficient methods for time series anomaly detection with the customer in the loop. The following are the current mainstream solutions to the issue of insufficient labels.

\textbf{Solution 1: unsupervised anomaly detector.}
Unsupervised Anomaly Detection (UAD) methods have been a major trend in dealing with unlabeled datasets~\cite{liu2008isolation, ren2019time, cleveland1990stl, guha2016robust, yue2022ts2vec}. These anomaly detectors use statistical methods, traditional machine learning methods, and deep learning methods to find anomalies in unlabeled data. However, it is difficult for a single UAD to achieve satisfying and robust results on all datasets. Namely, existing UAD methods are heuristic and not dataset-specific, thus \textbf{it is unclear which method is optimal for a specific dataset} without the guidance of supervision signals. Furthermore, UAD tends to have limited performance compared to supervised and semi-supervised methods. To this end, we need to develop a more effective and scalable anomaly detection method~\cite{ren2019time}.

\textbf{Solution 2: active learning.}
Active learning is a label-efficient supervised learning paradigm~\cite{settles2009active,schroder2020survey,cohn1994improving}. When given a large unlabeled data set, it iteratively selects a batch of data and queries their labels from oracle for downstream supervised learning. At each iteration, a model is trained with the current labeled set and used to select new data. This paradigm aims to render a trained model using a fraction of the data while achieving satisfactory performance.
However, the active learning paradigm often suffers from the so-called \emph{cold-start} problem~\cite{hacohen2022active}, \ie, \textbf{it requires a large initial set of labeled examples to work properly}~\cite{yuan-etal-2020-cold,pourahmadi2021simple}.
Nevertheless, in the field of time series anomaly detection, the active learning approach is not yet widespread~\cite{7929964}.

\textbf{Solution 3: weak supervision.}
Recently, the programmatic weak supervision paradigm~\cite{Ratner16,ratner2017snorkel,bach2019snorkel,zhang2022survey,zhang2021wrench} has been proposed to reduce manual efforts in training data labeling.
Instead of querying labels of each instance from human experts, programmatic weak supervision abstracts domain knowledge in the form of labeling functions (LFs). Each
LF inputs unlabeled data and outputs a pseudo label or treats it as abstain. Then, a \emph{label model}~\cite{Ratner16,Ratner19,fu2020fast} is used to aggregate the multiple potentially-noisy labels provided by LFs into a unique probabilistic training label for each instance.
The synthesized probabilistic training labels are in turn used to train a machine learning model.
As we can tell, the LFs play a core role in programmatic weak supervision, but \textbf{it is unclear how to generate LFs for time series data} since the raw values of time series are not informative enough for users to generate LFs accordingly.
In addition, to the best of our knowledge, we are the first that study the application of weak supervision methods on time series anomaly detection tasks.

Generally, none of the solutions could work well alone when facing the issue of insufficient labels in the overwhelming time series data. In this work, we propose a practical system seamlessly integrating these three mainstream solutions while at the same time overcoming their drawbacks. 

First of all, it is natural to leverage existing UADs for a new time series anomaly detection task, however, it is immensely difficult to figure the best UAD for the new task without an amount of manual effort, \eg, human annotation and evaluation.
To overcome this issue, in our system, \textbf{we leverage weak supervision method to integrate the outputs of multiple UADs instead of only applying one single UAD}, because the performance of such an ensemble strategy tends to be more robust across scenarios when we do not have a strong prior of which UAD works best. Specifically, each UAD is regarded as an independent LF and their votes are aggregated by a  weak supervision model of choice to produce the initial, potentially noisy labeling.

But the initial labels aggregated from multiple UADs are obtained in an unsupervised manner and therefore are application-agnostic and unlikely to work well for new scenarios, while active learning (AL) can help efficiently query application-specific labels from users.
And to improve the data efficiency of the labeling functions, we hope each of them holds as much unique information as possible. 
Then, \textbf{we incorporate active learning with a hybrid querying strategy in our system to interactively derive labeling functions for the most informative sample from users.}
In addition, \textbf{the unsupervised models serve as a warm-start to overcome the cold-start problem in the early stage of active learning}.
To the best of our knowledge, it is the first application of AL in labeling function creation.
In our design, WS and AL can promote each other in a reinforced manner rather than simply being added together. 
We have also designed an available real user interface, which facilitates the process of data labeling during the active learning process.

Finally, in active learning, compared to only querying labels for a single example each time, it is more sample-efficient to create an LF based on the queried label, which could provide extra information \cite{ratner2017snorkel}, such as patterns, heuristics, domain knowledge, \etc 
But it is difficult for customers to write heuristic labeling functions for a time-series-related task like other modalities such as natural language. Thus, \textbf{we propose a novel method that searches for similar timestamps based on contextual dense representations and utilizes the similarity search results as proxy labeling functions}. 
And these generated LFs, along with UAD-based ones, are further used to improve the involved weak supervision method.
This idea is also valuable in time-series-related tasks and many other domains.
Our primary \textbf{contributions} are summarized as follows: 
\begin{itemize}
    \item We propose LEIAD, a system that helps customers to promote their anomaly detection results in the industry via label-efficient interactions.
    We leverage programmatic weak supervision (PWS) for the first time and integrate the PWS with the active learning paradigm in an interactive system.
    With customers in the loop, the system effectively utilizes the limited supervision information provided by users to improve the accuracy of data labeling as well as the final performance of anomaly detection. 
    \item We incorporate both active learning and weak supervision in the system, enabling the mutual promotion of these two techniques for enhancing anomaly detection results. Specifically, we design an active learning strategy to generate more effective LFs, and leverage weak supervision models to produce more precise labeled data for training. 
    \item We introduce two methods to generate LFs for time series data. On the one hand, we treat each UAD as an initial LF to warmly start the active learning loop. On the other hand, we continuously generate LFs in the interactive process according to user annotated labels by searching similar time-series data points based on their dense representations. 
\end{itemize}
\section{Method}

\graphicspath{{images/}}

\subsection{System Overview}
\begin{figure}[htp]
\setlength{\abovecaptionskip}{-0.02cm}
 \setlength{\belowcaptionskip}{-0.5cm}
    \centering
    \includegraphics[width=7cm]{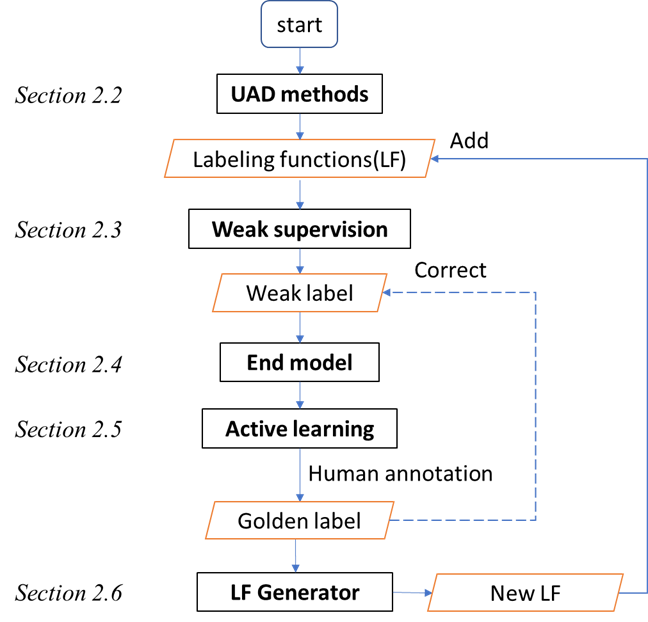}
    \caption{The overall pipeline of LEIAD.}
    \label{fig:pipeline}
\end{figure}

The whole pipeline of Label-Efficient Interactive Time-Series Anomaly Detection (LEIAD) is visualized in Figure \ref{fig:pipeline}. In the pipeline, we start from unsupervised anomaly detection methods, taking them as the initial set of labeling functions (LFs) and feeding them to a probabilistic weak supervision model. This weak supervision model aggregates the votes of multiple LFs to produce labels for training an end model. At each iteration $i$, an active learning agent is responsible for selecting the most informative points and visualizing the corresponding time-series segment for the end user to label. The user is asked to find out all label errors predicted by the previous end model inside that segment. In other words, both false positives and false negatives will be annotated, while other points will be associated with their predicted labels as ground truths. After receiving user feedback on that segment, we leverage a labeling function generator to extend the new information to a heuristic LF, append it to the set of existing LFs, and retrain the weak supervision model as well as the end model before starting iteration $i+1$.
The interaction loop continues until the end model is good enough as judged by some user-defined criteria, for example, the human annotator does not find error labels in the past 10 iterations.

In the following content, we first introduce unsupervised anomaly detectors as warm-start in Section \ref{sec:UAD}. Next, weak supervision methods and end models are introduced in Section \ref{sec:weak_upservision} and \ref{sec:end_model}, respectively. Afterwards, we present active learning in Section \ref{sec:active_learning} and LF generation in Section \ref{sec:label_function_generation}. 



\subsection{Unsupervised Anomaly Detector (UAD)}
\label{sec:UAD}
We adopt five common unsupervised anomaly detectors to construct the initial set of LFs, including i-Forest (Isolation Forest)~\cite{liu2008isolation, liu2012isolation}, SR (Spectral Residual)~\cite{ren2019time}, STL (Seasonal and Trend decomposition using Loess)~\cite{cleveland1990stl}, RC-Forest (Random Cut Forest)~\cite{guha2016robust}, and Luminol\footnote{https://github.com/linkedin/luminol}.


I-Forest builds an ensemble of binary isolated trees by iteratively selecting a feature and choosing a split value on the train set. In the isolate tree, the data point closer to the root has a higher probability of being an anomaly. SR first leverages Fourier Transform to filter out noises in the frequency domain and then performs anomaly detection based on the denoised saliency map. 
In STL, a time series is decomposed into trend, seasonal and reminder components. A threshold is set based on the remainder components to decide if a point is an anomaly. RC-forest is designed for data stream by dynamically maintaining data nodes on the tree.
Luminol is a lightweight Python library for time series data analysis released by Linkedin.

Since each UAD method has its own assumptions and privileges, an ensemble of them usually makes the anomaly detection results more robust. In this work, we will show that generating LFs with UAD and training an end model by weakly supervised weak labels can achieve further improvement compared to a simple ensemble of those UAD models. 
For a given UAD method, the corresponding UAD-based LF can be generated by the following steps. First, we infer all unlabeled training data, resulting in a predicted label $\hat{y}_i \in \{-1,0,1\}$ for each point $x_i$ in the time-series, where $-1$, $0$, and $1$ stand for abstain, normal, and anomalous points respectively. Then, we generate the corresponding LF, $\lambda_j(x_i)=\hat{y}_i$, for a specific UAD method $u_j$.


\subsection{Weak Supervision Module}
\label{sec:weak_upservision}
We leverage a weak supervision model to obtain weak labels for all timestamps from the votes of multiple LFs without any ground truth label.
This weak supervision model could be any of the existing label models in the literature, as long as it can effectively integrate the information brought by the LFs, such as  Snorkel \cite{ratner2017snorkel}, FlyingSquid \cite{fu2020fast}, and majority voting. 
A special note is that the weak supervision model inputs the votes of LFs and therefore is not applicable to new unseen data (\eg, test set) if the LFs cannot be applied; thus, it is typically used to provide supervision for training a classifier, which does not depend on LFs and can be readily generalized to new data~\cite{zhang2022survey}.

In our pipeline, we choose Snorkel \cite{ratner2017snorkel} for its good usability with the open source code\footnote{https://github.com/snorkel-team/snorkel} and robustness across various scenarios.
With Snorkel, we first apply all the LFs on the whole unlabeled training set, resulting in a weak label matrix $\Lambda$, where $\Lambda_{i,j}=\lambda_j(x_i)$ denotes the labeling result for the data point $x_i$ using the $j$-th LF. Then, a generative model $p_w(\Lambda, Y)$ can be defined as:
\begin{equation}
p_w(\Lambda, Y) = Z^{-1}_w{exp(\sum_{i=1}^m(w^T \Phi{_i}(\Lambda, y_i))} \\
\end{equation}
where $Y$ denotes the ground-truth label, which is unknown to the generative model; $w$ is the parameter for the model to optimize, $\Phi{_i}$ stands for the factor graph edge between weak label matrix $\Lambda$ and ground truth label matrix $y_i$, where $\Phi_{i,j}(\Lambda, Y) = \mathbf{1}\{\Lambda_{i,j}=y_i\}$; $Z^{-1}_w$ is the normalization factor. 
Since the ground truth label is unknown, Snorkel minimizes the negative log marginal likelihood of generative model over all possible ground truth set,
\begin{equation}
\hat{w} = \arg\min_{w} (-log \sum_Y p_w(\Lambda, Y))
\end{equation}
This objective can be optimized by interleaving stochastic gradient descent steps with Gibbs sampling \cite{hinton2002training}.

Then we construct the final weak label set $\hat L$ at those points with high confidence on Snorkel. We set two thresholds, $\tau_{pos}$ and $\tau_{neg}$, to separate normal and anomaly, according to preset the weak supervision ratio and anomaly percentage, while the others will be marked as abstain simultaneously. Additionally, weak labels will be corrected if they conflict with any known ground truths. Notably, we can 
utilize various data cleaning methods here, such as Cleanlab \cite{northcutt2021confidentlearning,northcutt2017rankpruning}, but such a simpler way is good enough to achieve effective results.


\subsection{End Model}
\label{sec:end_model}
\subsubsection{Models}
The end model for anomaly detection is trained on the final denoised weakly labeled training set $\hat L$, which contains only a most confident subset of the entire dataset for the current iteration, and is rendered to users as the output of our system. The end model can be any binary classification model that predicts whether a data point is anomalous based on the given feature set. The optimization objective can be formulated as:
\begin{equation}
\hat{\theta} = \arg \min_\theta \frac{1}{|\hat L|} \sum_{(x_i, y_i) \in \hat L} \ell(M_\theta(f(x_i)), y_i)
\end{equation}
where $\theta$ denotes the learnable parameter in the end model $M_\theta$, $f(\cdot)$ is the feature extractor, $\ell(\cdot)$ stands for the cross-entropy loss function, and $\hat{\theta}$ is the optimal parameter. In our implementation, we chose LightGBM as the end model, because of its relatively stable performance and competitive results in the experiments. 
LightGBM is based on GBDT (Gradient Boosting Decision Tree), which iteratively trains weak classifiers, \ie decision trees, to obtain the optimal model. LightGBM mainly introduces histogram optimization to improve the time and space efficiency of XGBoost without accuracy reduction, and thus becomes a popular model in major data competitions. 

\subsubsection{Features}
\label{sec:features}

Given a time-series instance $x_i$, the feature extractor $f(x_i)$ aims to output a collection of features. 
In this work, we adopt following features used in SR-DNN~\cite{ren2019time}, a state-of-the-art supervised model for univariate anomaly detection:

\begin{itemize}
    \item \textbf{Transformations.} We transform the value of each data point using logarithm function and take the result value as a feature.
    \item \textbf{Statistics.} We applied sliding windows to the time-series and treated the statistics calculated in each sliding window as features. The statistics we used include mean, exponential weighted mean, min, max, standard deviation, and the quantity of the data point values within a sliding window. We use multiple sizes of the sliding window to generate different features. The sizes are [10, 50, 100, 200, 500, 1440].
    \item \textbf{Ratios.} The ratios of current point value against other statistics or transformations
    \item \textbf{Differences.} The differences of current point value against other statistics or transformations.
\end{itemize}



\subsection{Active Learning}
\label{sec:active_learning}
Active learning is a paradigm in which, given a fixed budget of query, an agent aims to query labels for some selected data points from a user to maximize the performance of the end model. In our system, we introduce active learning to effectively and efficiently query labels from users, to remedy the initial labels aggregated from UADs. Such initial labels are obtained in an unsupervised manner and is unlikely to work well for each specific dataset/scenario. 
Specifically, we use the queried labels to replace the previous labels of those selected data and to generate new LFs to augment the initial set of LFs consisting of only UADs.
In each iteration, the active learning agent follows a certain pre-defined query strategy to select data points for users to label, while the goal is to improve the quality of the end model. In LEIAD, we use a hybrid query strategy consisting of two common used strategies in the field of active learning, namely, uncertainty and diversity, and one strategy designed specifically for the anomaly detection task, \textit{i.e.}, anomaly probabilities. They will be introduced separately in the following sub-sections. 


\subsubsection{Uncertainty}
\begin{enumerate}
    \item \textbf{Agreement of labeling functions.} Suppose we have $M$ labeling functions $\{LF_1, LF_2, ..., LF_M\}$; each of them gives a label $y^M_{i,t} \in \{0, 1\}$ to the corresponding timestamp. The probabilities of a given timestamp to be anomaly can be calculated by voting of all LFs. 
    \begin{align}
    p^u(x_{i,t}) = \frac{1}{M} \sum_{i=1}^{M} y^M_{i,t}
    \end{align}
    Then, we leverage cross-entropy to measure the agreement of LFs. 
    \begin{align}
    A(x_{i,t}) = p^u(x_{i,t}) \log(p^u(x_{i,t})) + (1 - p^u(x_{i,t}))\log(1-p^u(x_{i,t})) 
    \end{align}
    A timestamp should be sampled with larger probability if the agreement at that point is smaller (the cross-entropy metric is larger).
    \item \textbf{Abstention of labeling functions.}
    \begin{align}
    H(x_{i,t}) = \log{(count - count(x_{i,t}) + 1)}
    \end{align}
    $count$ is the number of LFs, $count(x_{i,t})$ is the count of LFs that give the corresponding timestamp a pseudo label (other than \textit{Unknown}). The sampling ratio should be larger if this count is small. $\log$ makes a flatter distribution.
    \item \textbf{Uncertainty of the supervised end model}
    The mutual information is adopted to reflect the uncertainty of supervised end model. Assume the probability given by the supervised end model is $p^s$. For a certain timestamp $x_{i,t}$, the metrics is calculated as follows. 
     \begin{align}
    U(x_{i,t}) = p^s(x_{i,t})\log(p^s(x_{i,t})) + (1 - p^s(x_{i,t}))\log(1-p^s(x_{i,t})) 
    \end{align}
    The larger mutual information, the larger the sampling ratio in the query strategy.
\end{enumerate}

\subsubsection{Diversity}
In addition to model certainty, we also consider diversity in the sample selection procedure. In other words, if similar sample has been labeled in former samples, then the query probability should become lower. 
\begin{align}
& sim(x_i, x_j) = r(x_i) \cdot r(x_j) \\
& D(x_{i,t}) = 1 - \frac{1}{|S^L|}\sum_{x_k \in S^L}sim(x_{i,t}, x_k)
\end{align}
where $r(.)$ denotes the contextual representation derived by TS2Vec ~\cite{yue2022ts2vec}. $S^L$ is the labeled set of timestamps collected by previous interactions, and $|S^L|$ is the number of labeled timestamps. Note that we should never select a timestamp if it has already been labeled, \textit{i.e.}, $x_{i,t} \in S^L$.

\subsubsection{Anomalous Probability}
We add anomalous probability item to choose anomalous points which are thought more informative. The probability is calculated using the average of UAD scores $\{UAD_1, UAD_2, ..., UAD_N\}$.

\begin{align}
    P(x_{i,t}) = \frac{1}{N} \sum_{i=1}^{N} p^N_{i,t}
\end{align}

\subsubsection{Query Function}
Finally, we combine the metrics above to formulate a hybrid query strategy. 
\begin{align}
Q(x_{i,t}) = A(x_{i,t}) + \alpha H(x_{i,t}) + \beta U(x_{i,t}) + \gamma D(x_{i,t}) + \delta P(x_{i,t})
\end{align}
where $A(\cdot)$, $H(\cdot)$, $U(\cdot)$, $D(\cdot)$ and $P(\cdot)$ stand for \textit{agreement of LFs}, \textit{abstention of LFs}, \textit{uncertainty of the supervised end model}, \textit{diversity} and \textit{anomaly probability}, respectively. 

\subsection{Labeling Function Generation}
\label{sec:label_function_generation}
Finally, we present how to generate new LFs based on the queried labels and similarities among data points.
In our system, we utilize TS2Vec~\cite{yue2022ts2vec}, a universal framework to obtain time-series representation based on self-supervised losses, to derive $k$-dimensional dense representation vectors for all points in the time series data. 
Given a user annotation $(x_i, y_i)$ in the new interaction round, we generate a new LF by querying similar points and expanding the same label to those selected points.
Specifically, We adopt 
$L_1$ distance as measure of similarity:
\begin{align}
& sim_{l1}(x_i, x_q) = \parallel r(x_i) - r(x_q)\parallel_1
\end{align}
where $x_q$ represents the candidate points to be selected. According to the degree of deviation from the mean, we get a candidate set
$P_{l1}$, 
the candidate set will be labeled as $y_i$, while other points will be labeled as abstain. 
\begin{align}
& \lambda(x_i) = 
    \begin {cases}
    y_i  &\text{if } x_i \in P_{l1} \\ 
    -1 &\text{others}
    \end{cases}\\
& P_{l1} = \{ x \in S \mid sim_{l1}(x, x_q) < \mu_{l1} - \tau_{l1}\sigma_{l1} \} 
\end{align}
where $\mu$ and $\sigma$ correspond to means and standard deviations, while $\tau$ is a hyper-parameter. 
At the same time, we also provide an optional statistic model leveraging statistics features to measure similarity, which means replacing the embeddings with vectors containing various statistics in windows of different sizes. 

In large datasets like KPI, to strike a balance between performance and runtime, we use ScaNN \cite{guo2020accelerating} to search for similar vectors efficiently, which can accelerate large-scale inference of maximum inner product searching (MIPS) procedure. In our settings, we implement partitioning before scoring, and rescoring are adopted to enhance accuracy.

\section{Experiment}
\subsection{Experimental Setup}
\subsubsection{Testbed}

We conducted our experiments on Ubuntu 18.04 LTS. The machine consists an Intel(R) Xeon(R) E5-2690 v3 CPU, 224GB DRAM, and 4 Nvidia Tesla M60 GPUs.

\subsubsection{Datasets}

We simulated our system to three uni-variate anomaly detection datasets. The role of human annotators is to provide labels for the timestamps selected by the system.
The purpose of simulation experiments is to evaluate the system in more scenarios without too much human cost, while other interaction designs keep the same as the real user study in Section 4.

\textbf{Yahoo}\footnote{https://yahooresearch.tumblr.com/post/114590420346/a-benchmark-dataset-for-time-series-anomaly} is a small dataset which contains 367 lines and about 580,000 points in total, and the anomaly ratio is around 0.6 percent. 

\textbf{KPI}\footnote{https://github.com/NetManAIOps/KPI-Anomaly-Detection} is a large dataset which has 6,000,000 points, and the anomaly ratio is about 2.5 percent. 

\textbf{Microsoft} is an internal production dataset, which has about 300,000 points in 1496 lines, and the anomaly ratio is around 8 percent. 

For Yahoo and Microsoft dataset collected by a commercial anomaly detection system, we split the train and test data randomly by 3:1. For KPI dataset, we adopt the original split, which is about 1:1 for train/test set.

\subsubsection{Metrics}
We use two general metrics (i.e., AP and ROC AUC) to measure the performance. iteration 

\textbf{Average Precision (AP)} summarizes a precision-recall curve as the weighted mean of precisions achieved at each threshold, with the increase in recall from the previous threshold used as the weight.

A receiver operating characteristic (ROC) curve is a graph showing the performance of a classification model at all classification thresholds.
\textbf{Area Under the ROC Curve (ROC AUC)} provides an aggregate measure of performance across all possible classification thresholds. But it may be affected by a large amount of true negative samples due to the inherent bias of the AD task.

\subsubsection{Compared Methods}
We run our algorithms on three datasets and exhibit the performance of the 5th, 10th, 20th, 50th and 100th iterations for comparison. The end model we use in all experiments is the same, which is LightGBM with various time-series features as described in section \ref{sec:features}.
\begin{itemize}
    \item \textbf{Snuba}~\cite{varma2018snuba} is a weak supervision system that automatically generates labeling functions via tree-based models using a small labeled dataset. Based on the labeling functions, the system assigns training labels to a large unlabeled dataset, which is used to train the end model for anomaly detection. 
    \item \textbf{Random AL} randomly selects time-series segments for annotation. In each iteration, it randomly samples a point from unlabeled dataset and presents the corresponding time-series segment to the human annotator. After receiving a new labeled instance, the end anomaly detection model will be re-trained by adding this data to the training set. 
    \item \textbf{Uncertainty-based AL} is an active learning baseline, which queries informative instances with uncertainty sampling. We implement the Uncertainty-based AL Algorithm with modAL \cite{danka2018modal} library.
    \item \textbf{Hybrid AL} is the active learning method based on the hybrid query strategy proposed in this paper. Its difference with full LEIAD is that it only leverages ground truth labels for training, but does not utilize weak supervision models to obtain weak labels as data augmentation.
    \item \textbf{LEIAD} is the method proposed in this paper. It provides a user-friendly interface for customers in-the-loop to efficiently promote the time-series anomaly detection performance for different scenarios. LEIAD integrates multiple techniques in the interactive framework, including unsupervised anomaly detectors, LF-based active learning, weak supervision and annotation-based labeling function generator, which will be further analyzed individually in our analyses.
\end{itemize}

\subsubsection{Parameters setting}
\label{sec:param_set}


In Table \ref{table:parameters_setting} we present core hyper-parameters and their value sets in our experiments. The \textit{anomaly percentage} is the percentage of anomalies in the dataset estimated by the user. The \textit{weak supervision ratio} indicates the percentage of weak labels in the total data samples in the initial phase. As queried labels increase, we set the number of weak labels as double of the queried labels. The \textit{length of segment} is the length of time-series segment shown to the user. The \textit{number of neighbors} refers to the number of neighbors searched by ScaNN. 

\begin{table}[h]
\setlength{\abovecaptionskip}{-0.03cm}
 \setlength{\belowcaptionskip}{-0.4cm}
\caption{Parameters setting based on dataset}
\label{table:parameters_setting}
\begin{tabular}{llll}
\hline
Parameter              & Microsoft & Yahoo & KPI   \\ \hline
Anomaly percentage     & 5         & 1     & 2     \\
Weak supervision ratio & 0.1       & 0.1   & 0.05  \\
Length of segment      & 100        & 400   & 800   \\
Number of neighbors        & 50        & 200   & 1000  \\ \hline
\end{tabular}
\end{table}

\subsection{Overall Results}


As in Table \ref{table:overall_results}, \textit{LEIAD} renders superior performance across diverse datasets. In the first 20 iterations, we achieved significant improvement compared with almost all baselines. From the results, we can see that the performance of \textit{Snuba} is relatively unstable (\ie, the standard deviation is higher than other methods). We speculate that this is because \textit{Snuba} does not involve an active learning module to query the oracle while instead uses randomly sampled data and leading to higher randomness. Considering AL baselines, \textit{Random AL} may be limited by the retrieval of duplicated information, it suffers from worse sample efficiency as a large amount of time-series segments have been well covered by unsupervised anomaly detectors while adding annotations on them brings little extra benefits.
However, active learning is a promising protocol to accelerate this process but sometimes causes a biased distribution of the derived training sets. As shown in Table \ref{table:overall_results}, \textit{Uncertainty-based AL} achieves significant advantages in the early stage but becomes inferior to \textit{Random AL} in later stages. Intuitively, using a diversity metric may mitigate the distribution bias; meanwhile, labeling functions naturally contain much uncertain information. Thus, we leverage a novel hybrid query strategy to improve the performance of active learning (Refer to \textit{Hybrid AL}).
Moreover, for the comparison with the best unsupervised anomaly detector, LEIAD performs worse at the beginning, while after sufficient interactions, the performance of LEIAD can consistently beat other state-of-the-art methods.

The proposed system \textit{LEIAD}, equipped with Hybrid AL and weak supervision jointly,
can efficiently query user feedback and extract effective information from noisy LFs and outperforming almost all baselines significantly. Specifically, in Microsoft dataset, \textit{LEIAD} exceeds \textit{Random AL} by almost 50\% in the early stages and beats Uncertainty-based AL by 10-20\%. Meanwhile, our \textit{Hybrid AL} method achieves at least 10\% better final performance than other baselines as evaluated by AP metrics. For AUC ROC metrics, \textit{LEIAD} outperforms \textit{Random AL} and \textit{Uncertainty-based AL} by 5-15\% in the first 20 iterations on Microsoft and Yahoo datasets. On most occasions, \textit{LEIAD} achieves more stable results than other baselines,
the same conclusion can be drawn for the AUC ROC metrics.

\begin{table*}[]
 \setlength{\abovecaptionskip}{-0.02cm}
 \setlength{\belowcaptionskip}{-0.2cm}
\caption{Overall results.}
\label{table:overall_results}
\scalebox{0.85}{
\begin{tabular}{clcccccccccc}
\hline
\multirow{3}{*}{Dataset}   & \multicolumn{1}{c}{\multirow{3}{*}{Compared Method}} & \multicolumn{5}{c}{AP}                                                                                                                                                                                     & \multicolumn{5}{c}{ROC   AUC}                                                                                                                                                                              \\ \cline{3-12} 
                           & \multicolumn{1}{c}{}                                 & \multicolumn{5}{c}{Iteration}                                                                                                                                                                              & \multicolumn{5}{c}{Iteration}                                                                                                                                                                              \\ \cline{3-12} 
                           & \multicolumn{1}{c}{}                                 & 5                                      & 10                                     & 20                                     & 50                                     & 100                                    & 5                                      & 10                                     & 20                                     & 50                                     & 100                                    \\ \hline
\multirow{6}{*}{Yahoo}     & Best UAD                                             & \multicolumn{5}{c}{0.27}                                                                                                                                                                                   & \multicolumn{5}{c}{0.85}                                                                                                                                                                                   \\
                           & Snuba                                                & 0.12±0.10                              & 0.12±0.09                              & 0.22±0.10                              & 0.23±0.11                              & 0.26±0.10                              & 0.70±0.13                              & 0.69±0.15                              & 0.73±0.09                              & 0.73±0.09                              & 0.74±0.05                              \\
                           & Random AL                                            & 0.11±0.08                              & 0.11±0.05                              & 0.15±0.08                              & 0.23±0.06                              & 0.27±0.09                              & 0.69±0.10                              & 0.72±0.09                              & 0.69±0.10                              & 0.76±0.06                              & 0.77±0.06                              \\
                           & Uncertainty-based AL                                 & 0.16±0.11                              & 0.19±0.10                              & 0.17±0.05                              & 0.33±0.09                              & 0.53±0.03                              & 0.80±0.08                              & 0.75±0.14                              & 0.89±0.02                              & \textbf{0.93±0.02}                     & \textbf{0.96±0.01}                     \\
                           & Hybrid AL                                            & 0.33±0.10                              & \textbf{0.44±0.06}                     & 0.46±0.04                              & 0.53±0.05                              & \textbf{0.59±0.03}                     & 0.84±0.09                              & 0.86±0.02                              & 0.88±0.04                              & 0.89±0.04                              & 0.92±0.03                              \\
                           & LEIAD                                                & \textbf{0.36±0.06}                     & 0.44±0.04                              & \textbf{0.50±0.02}                     & \textbf{0.56±0.03}                     & 0.58±0.04                              & \textbf{0.89±0.02}                     & \textbf{0.87±0.02}                     & \textbf{0.89±0.04}                     & 0.88±0.04                              & 0.91±0.04                              \\ \hline
\multirow{6}{*}{KPI}       & Best UAD                                             & \multicolumn{5}{c}{0.26}                                                                                                                                                                                   & \multicolumn{5}{c}{0.76}                                                                                                                                                                                   \\
                           & Snuba                                                & 0.05±0.04                              & 0.12±0.08                              & 0.07±0.08                              & 0.07±0.03                              & 0.08±0.05                              & 0.58±0.11                              & 0.69±0.10                              & 0.51±0.11                              & 0.61±0.10                              & 0.66±0.08                              \\
                           & Random AL                                            & 0.06±0.04                              & 0.11±0.09                              & 0.13±0.07                              & 0.16±0.09                              & 0.22±0.07                              & 0.56±0.11                              & 0.61±0.19                              & 0.68±0.10                              & 0.75±0.04                              & 0.80±0.04                              \\
                           & Uncertainty-based AL                                 & \multicolumn{1}{l}{0.12±0.01}          & \multicolumn{1}{l}{0.14±0.05}          & \multicolumn{1}{l}{0.14±0.03}          & \multicolumn{1}{l}{0.19±0.03}          & \multicolumn{1}{l}{0.25±0.04}          & \multicolumn{1}{l}{\textbf{0.80±0.02}} & \multicolumn{1}{l}{\textbf{0.83±0.04}} & \multicolumn{1}{l}{0.80±0.03}          & \multicolumn{1}{l}{0.83±0.02}          & \multicolumn{1}{l}{0.85±0.01}          \\
                           & Hybrid AL                                            & \multicolumn{1}{l}{0.09±0.04}          & \multicolumn{1}{l}{0.10±0.02}          & \multicolumn{1}{l}{0.18±0.06}          & \multicolumn{1}{l}{0.28±0.04}          & \multicolumn{1}{l}{\textbf{0.40±0.04}} & \multicolumn{1}{l}{0.68±0.09}          & \multicolumn{1}{l}{0.71±0.07}          & \multicolumn{1}{l}{0.75±0.04}          & \multicolumn{1}{l}{0.83±0.01}          & \multicolumn{1}{l}{0.87±0.02}          \\
                           & LEIAD                                                & \multicolumn{1}{l}{\textbf{0.20±0.03}} & \multicolumn{1}{l}{\textbf{0.20±0.01}} & \multicolumn{1}{l}{\textbf{0.23±0.05}} & \multicolumn{1}{l}{\textbf{0.29±0.05}} & \multicolumn{1}{l}{0.39±0.03}          & \multicolumn{1}{l}{0.70±0.07}          & \multicolumn{1}{l}{0.72±0.08}          & \multicolumn{1}{l}{\textbf{0.82±0.03}} & \multicolumn{1}{l}{\textbf{0.84±0.02}} & \multicolumn{1}{l}{\textbf{0.87±0.01}} \\ \hline
\multirow{6}{*}{Microsoft} & Best UAD                                             & \multicolumn{5}{c}{0.25}                                                                                                                                                                                   & \multicolumn{5}{c}{0.71}                                                                                                                                                                                   \\
                           & Snuba                                                & 0.11±0.03                              & 0.14±0.03                              & 0.13±0.03                              & 0.10±0.02                              & 0.09±0.01                              & 0.70±0.13                              & 0.69±0.15                              & 0.73±0.09                              & 0.73±0.09                              & 0.74±0.05                              \\
                           & Random AL                                            & 0.18±0.05                              & 0.21±0.04                              & 0.25±0.04                              & 0.34±0.02                              & 0.40±0.02                              & 0.64±0.05                              & 0.67±0.04                              & 0.71±0.03                              & 0.76±0.02                              & 0.79±0.01                              \\
                           & Uncertainty-based AL                                 & 0.19±0.07                              & 0.28±0.02                              & 0.31±0.02                              & 0.35±0.02                              & 0.38±0.01                              & 0.64±0.06                              & 0.70±0.02                              & 0.73±0.01                              & 0.76±0.01                              & 0.77±0.00                              \\
                           & Hybrid AL                                            & 0.21±0.03                              & 0.22±0.05                              & 0.30±0.02                              & 0.38±0.01                              & \textbf{0.44±0.00}                     & 0.66±0.03                              & 0.67±0.03                              & 0.72±0.02                              & 0.77±0.01                              & 0.80±0.01                              \\
                           & LEIAD                                                & \textbf{0.30±0.01}                     & \textbf{0.32±0.01}                     & \textbf{0.35±0.01}                     & \textbf{0.39±0.00}                     & 0.43±0.01                              & \textbf{0.73±0.01}                     & \textbf{0.75±0.01}                     & \textbf{0.76±0.01}                     & \textbf{0.78±0.00}                     & \textbf{0.81±0.01}                     \\ \hline
\end{tabular}
}
\end{table*}




\subsection{Analysis}
\label{sec:analysis}

\subsubsection{The effect of aggregating multiple UADs as warm-up.}
\begin{figure*}[htp]
\setlength{\abovecaptionskip}{-0.02cm}
 \setlength{\belowcaptionskip}{-0.4cm}
\begin{minipage}{0.3\textwidth}
    \raggedleft
    \includegraphics[width=\textwidth]{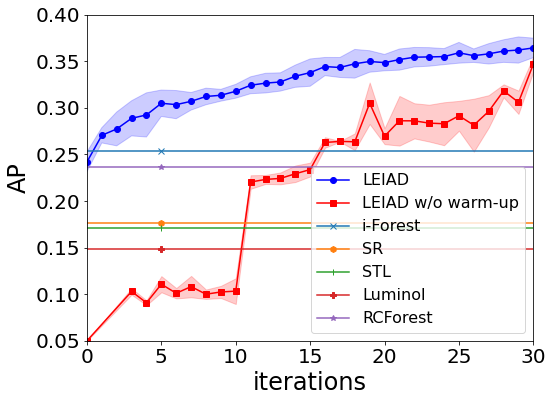}
    \caption{Average precision curve with v.s. without warm-up, and UADs are plotted as reference lines.}
    \label{fig:warm_up}
\end{minipage}
\hfill
\begin{minipage}{0.3\textwidth}
    \centering
    \includegraphics[width=\textwidth]{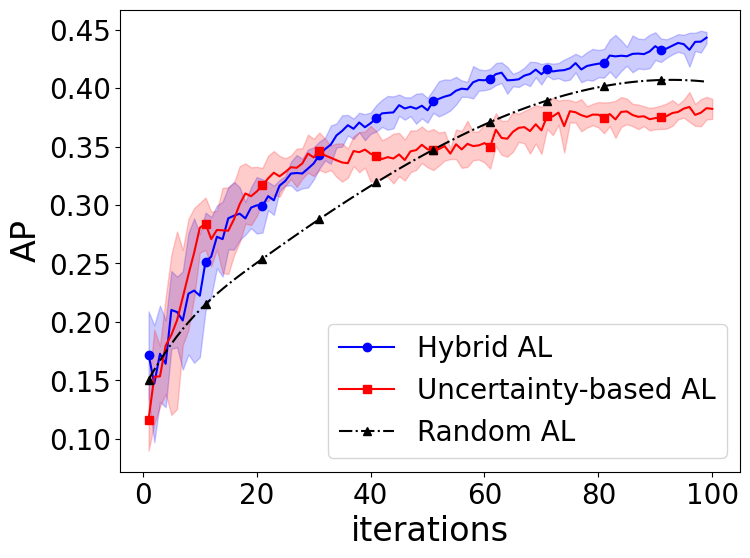}
    \caption{Different active learning in Microsoft dataset.}
    \label{fig:kensho_al}
\end{minipage}
\hfill
\begin{minipage}{0.3\textwidth}
    \raggedright
    \includegraphics[width=\textwidth]{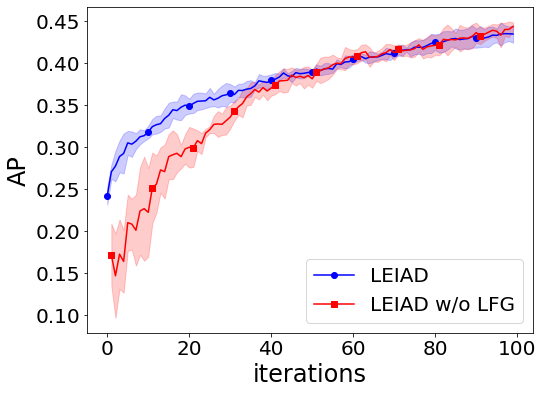}
    \caption{Ablation of labeling function generator on Microsoft dataset.}
    \label{fig:kensho_lfg}
\end{minipage}
\end{figure*}

In our preliminary studies, we observe that the performance of different UADs varies dramatically across datasets, and there is no such an always-winner.
Therefore, we apply the weak supervision method to aggregate the outputs of multiple UADs as initial weak labels and warm-up for the later stage of active learning.
Such an ensemble technique could render robust performance across datasets and performs slightly worse than the best UAD (i-Forest in Figure~\ref{fig:warm_up}) at first. Furthermore, from Figure~\ref{fig:warm_up} we can see that the weak supervision method can readily outperform the best UAD after only one iteration of active learning with new LF added. This observation verifies the superiority of aggregating multiple UADs compared to using only one UAD.

We are also curious about the effect of the warm-up for handling cold-start issues in active learning.
Thus, we compare our system against its counterpart without warm-up in Figure~\ref{fig:warm_up}. The results show that our system with warm-up is consistently better than that without warm-up and has a more smooth AP curve. That is because, without warm-up, active learning is vulnerable to distribution bias and label imbalance in the initial stage. We also observe that the warm-up with UADs tends to produce a similar label distribution to the ground truth and is likely to make our system robust to the sample bias at the beginning, making the trained model more stable across iterations.


\subsubsection{The effect of active learning and the proposed hybrid strategy.}
In this section, we first study the effect of incorporating active learning (AL) in our system.
In Figure \ref{fig:kensho_al}, we present the performance curves for using three different strategies of active learning: random, uncertainty, and hybrid.
From the results, we can see that our hybrid strategy only requires 50\% of interactions to achieve the same performance as the random strategy, and provides a 20-30\% performance gain over the random strategy at the same number of interactions. The comparison demonstrates the effectiveness of active learning for our system since the random strategy can be regarded as a manual annotation process without active learning.
Compared with the uncertainty strategy, a frequently used baseline of active learning, our hybrid strategy obtains a maximum of 10\% performance gain, and the performance gap tends to increase with more iterations.
This result demonstrates the superiority of the proposed hybrid strategy that takes more information into consideration (\eg, the agreement among LFs) than the uncertainty strategy.

\subsubsection{The effect of labeling function generator.}
Here, to study the effect of the labeling function generator, we present the performance curves of our system and its variance without labeling function generator in Figure \ref{fig:kensho_lfg}.
From the result, we can see that adding new LFs leads to superior performance at the early stage of active learning without extra manual effort.
Although both curves achieve similar AP scores at the later stage, the labeling function generator could improve the sample efficiency at the beginning and therefore is beneficial when users have a limited annotation budget.



\subsubsection{The effect of ScaNN on accelerating similarity search.}
\begin{figure}[htp]
\setlength{\abovecaptionskip}{-0.02cm}
 \setlength{\belowcaptionskip}{-0.4cm}
    \centering
    \includegraphics[width=6cm]{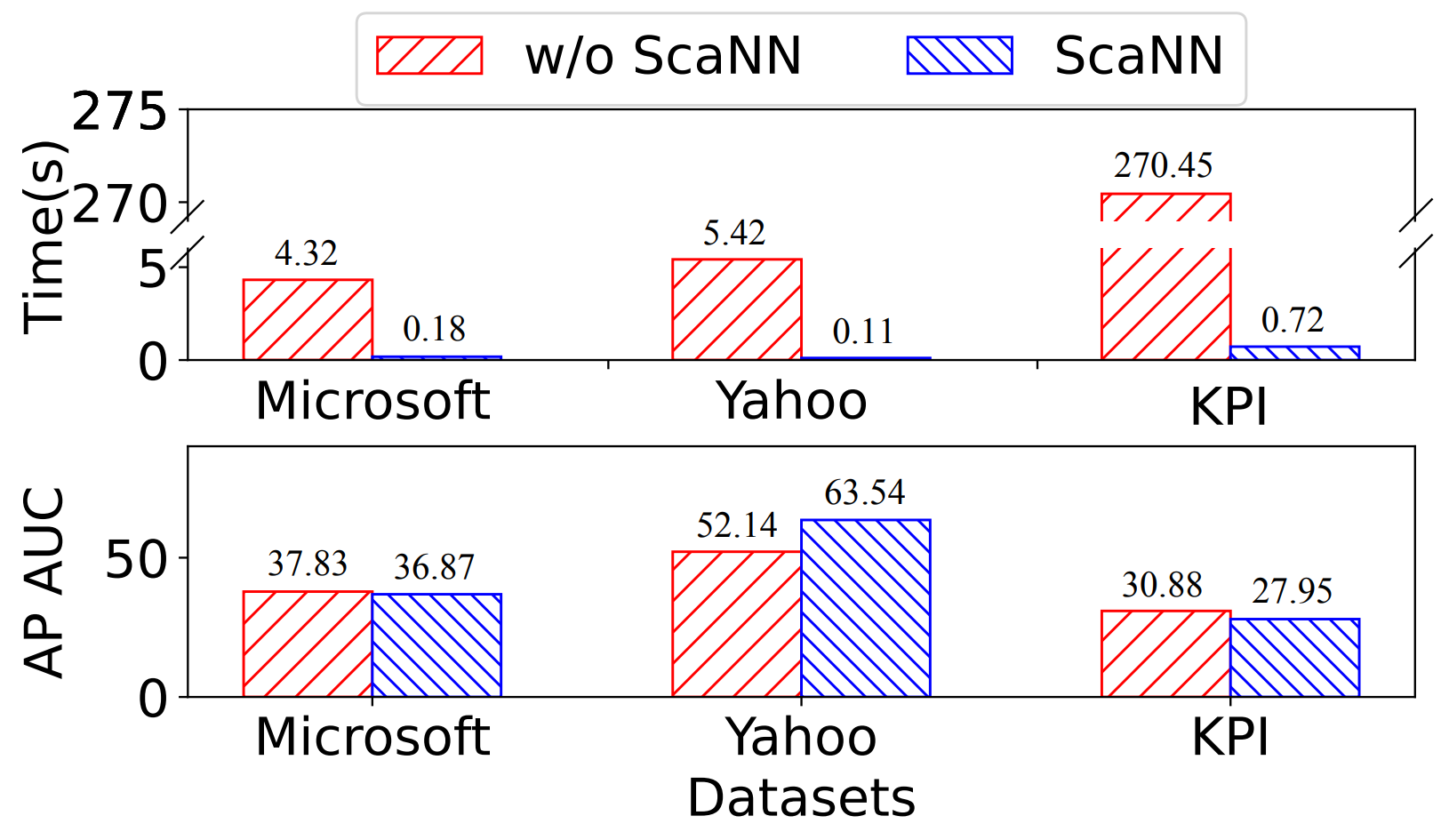}
    \caption{The above graph shows the average running time(s) of each iteration on Microsoft, Yahoo and KPI respectively, the below are their AP AUC, which is the area under their AP curves.}
    \label{fig:scann}
\end{figure}

The similarity search step in the labeling function generator is the major bottleneck of runtime for large datasets like KPI. We exploit ScaNN \cite{guo2020accelerating} to accelerate the similarity search.
As shown in Fig \ref{fig:scann}, ScaNN reduces the search time by more than 95\%, while only causing a slight performance drop in the Microsoft and KPI dataset. While in the Yahoo dataset, ScaNN even achieves better performance.
Here, we use Area Under the AP Curve (AP AUC) to evaluate the overall performance of the model, which is a composite evaluation metric considering the curve growth trend. 

\section{User Study}

\begin{figure}[htp]
\setlength{\abovecaptionskip}{-0.02cm}
 \setlength{\belowcaptionskip}{-0.9cm}
    \centering
    \includegraphics[width=7cm]{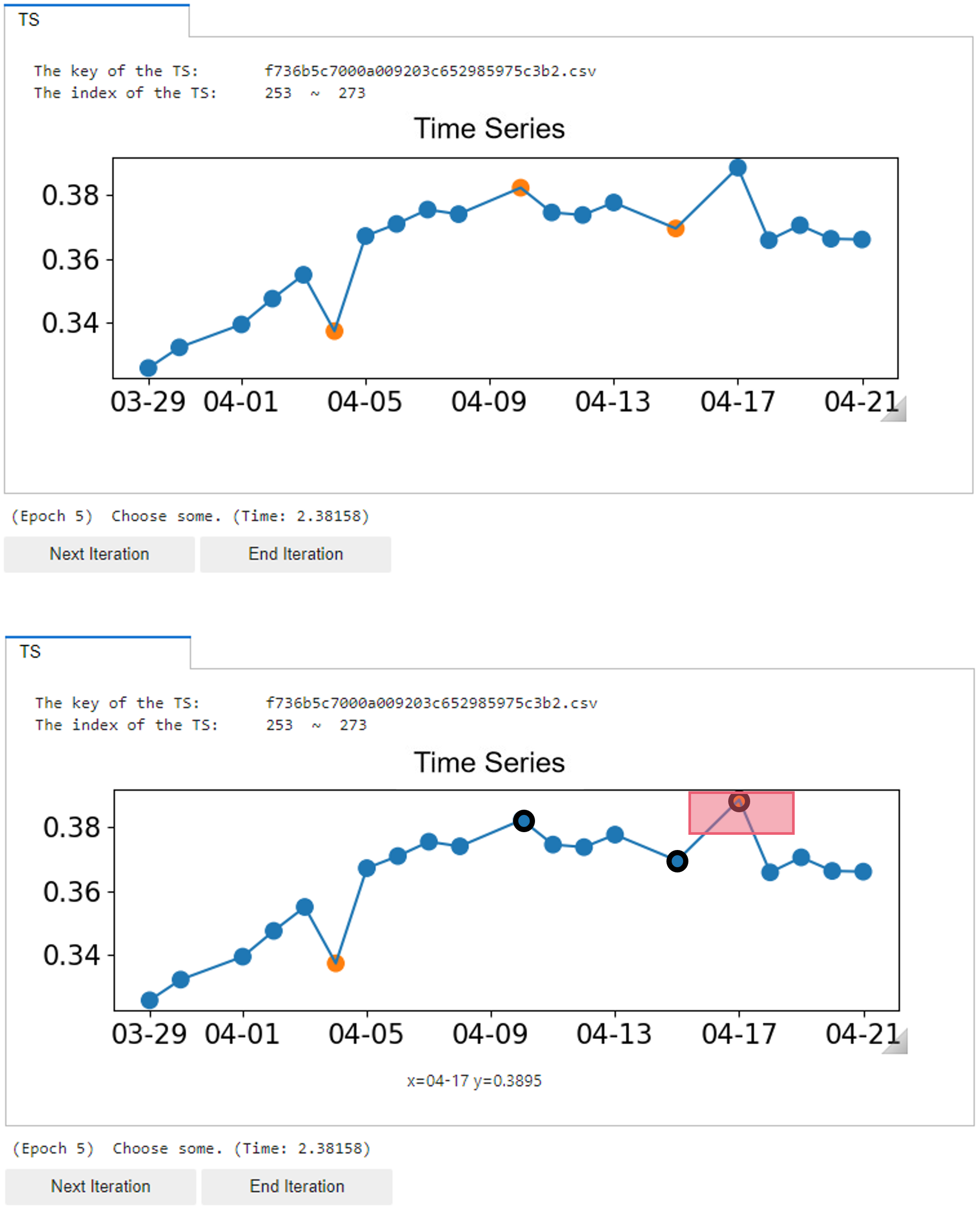}
    \caption{LEIAD demo interface. The above figure shows the user interface, and the prediction result of the previous iteration has been marked in orange. The figure below shows the process of the user correcting the label, and the corrected data points are bolded in black.}
    \label{fig:interface}
\end{figure}

\begin{figure}[htp]
\setlength{\abovecaptionskip}{-0.02cm}
 \setlength{\belowcaptionskip}{-0.6cm}
    \centering
    \includegraphics[width=7cm]{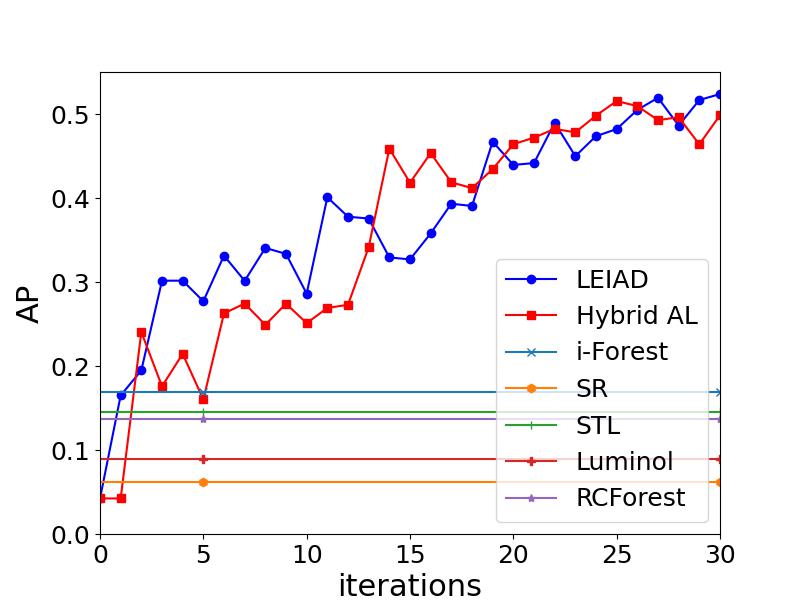}
    \caption{The performance of LEIAD on real dataset.}
    \label{fig:user_study}
\end{figure}

We invite a team within the organization to evaluate the performance of the system. The team is building mobile applications and they are monitoring key shipping metrics to auto-detect abnormal movements. SSR, TTS, PCR, CCR and coverage metrics are included for 18 end points including main clients. There are 861 series in total. We build an interactive interface with the pipeline shown in Figure~\ref{fig:pipeline}. The process can be found in Figure~\ref{fig:interface}. Specifically, for each iteration, a time series will be shown to the user. The user needs to annotate anomalies within the time range and submit them. In the next iteration, a selected time series with highlighted anomalies will be presented to the user. The anomalies are detected based on the user's previous annotations. The user will label as many iterations as they can. The evaluation will follow this method: assuming the user labels $n$ iterations, for each model learned in the $i_{th}$ iteration, we calculate average precision based on the ground truth collected from the ${i+1}_{th}$ iteration to the $n_{th}$. The objective of the study can be summarized as follows: 
 \begin{itemize}
     \item How real users interact with the system and whether the process is convenient for the user to provide feedback.
     \item How many iterations need to be taken before a user get satisfying results with the whole dataset.
 \end{itemize}
Here we suggest the customer to refer to the evaluation metrics provided by the system after some rounds of annotations. The curve of metrics has a trend of convergence, which will help the annotator determine whether to stop labeling. 

In Figure~\ref{fig:user_study}, the average precision of this real dataset is converged after 25 iterations. The converged average precision is about 5\% higher than the supervised model trained only with ground truth annotations and 200\% higher than the unsupervised model. 
\section{Related work}
We discuss related studies with respect to the following three aspects.

\textbf{Unsupervised Anomaly Detection.}
There has been a long history of unsupervised anomaly detection on time series. A general and robust work is the STL \cite{cleveland1990stl} algorithm proposed by Cleveland, which mainly uses locally weighted scatterplot smoothing to fit curves. Isolation forest \cite{liu2008isolation, liu2012isolation}, also known as iForest, is another famous machine learning algorithm, exploiting an isolation mechanism to detect anomalies. Following the lens of random cut forests, RRCF \cite{guha2016robust} explored a randomized approach to improve the performance of iForest for input streams. Recently, more researchers strive to combine methods in CV and NLP domain with time series analysis. Inspired by the Spectral Residual (SR) model from visual saliency detection domain, Ren et al. \cite{ren2019time} innovatively combined SR and CNN together in time series. TS2Vec \cite{yue2022ts2vec} provided a universal method to learn unsupervised time series representations via hierarchical contrastive learning, effectively helping downstream anomaly detection tasks.

\textbf{Active Learning.}
There have been some studies trying to overcome the difficulties of active learning application in time series anomaly detection with promising results. ACTS \cite{7929964} has proposed two informativeness metrics in time series, namely uncertainty and utility. Our active learning strategy not only considers the informativeness of the time series itself, but also incorporates the informativeness brought by LFs. Subsequently, \cite{trittenbach2021overview} found that there is no one perfect universal active learning strategy that can be applied to all scenarios  after comparing several methods. Recently, some novel usages of active learning emerged. For instance, RLAD \cite{wu2021rlad} proposed a novel approach that combines active learning and reinforcement learning for anomaly detection. 

\textbf{Weak Supervision.} 
Since LFs are difficult to be obtained for time series, there are few related works on weak supervision of time series. We introduce some general weak supervision frameworks incorporating active learning, which are similar to our work. Nashaat et al. \cite{nashaat2018hybridization} has used active learning to correct the output of weak supervision to improve the performance of the model. Active WeaSuL \cite{biegel2021active} proposes improvements on the basis of \cite{nashaat2018hybridization}, improving the query strategy and the loss function of the weak supervision model. IWS \cite{boecking2020interactive} adopts interactive weak supervision, in which the model automatically generates and displays a large number of LFs, while the user only needs to judge whether the LFs are stronger than random classifiers. In our work, active learning helps to select the proper unlabeled points to generate heuristics, which distinguishes it from previous work.
\section{Conclusion}
In this paper, we propose a new system, LEIAD, for label-efficient interactive time-series anomaly detection. Using the system, customers can start from unsupervised anomaly detection experiences and improve the quality of anomaly detection efficiently by playing with the system for a few interactions. 
In this respect, LEIAD wisely encapsulates active learning, labeling function (LF) generator and weak supervision. 
It interactively collects feedbacks from users based on active learning, automatically generates new labeling functions by the LF generator, and constructing more and more accurate labeled dataset via weak supervision to improve the performance of target anomaly detector.
Empirically, the proposed system achieved satisfactory performance, outperforming previous methods when simulating on three time-series anomaly detection datasets. User study in a real-world scenario also demonstrated good feasibility of the system.



\bibliographystyle{ACM-Reference-Format}
\bibliography{sample-base}

\newpage
\appendix

\section{appendix}

As shown in Figure \ref{fig:pipeline} , our method is composed of five components (i.e., an unsupervised anomaly detector, a weak supervision model, a labeling function generator, an active learning model and an end model). The parameters used in these models are summarized in Table \ref{table:parameters_setting_1}. 

\begin{table}[t]
\caption{Hyper-parameter settings}
\label{table:parameters_setting_1}
\begin{tabular}{lll}
\hline
                & Hyper-parameter Name                          & Value          \\ \hline
i-Forest        & number of estimators       & 1000           \\
                & contamination              & 0.01           \\ \hline
STL             & period                     & 90             \\
                & lo\_frac                   & 0.60           \\
                & lo\_delta                  & 0.01           \\ \hline
RCForest        & shingle size               & 1              \\
                & number of trees            & 100            \\ \hline
SR              & mag window                 & 200            \\
                & score window               & 10             \\ \hline
Snorkel         & learning rate              & 0.001          \\
                & training epoch             & 200            \\ \hline
TS2Vec          & is\_normalized             & True           \\
                & similarity measurement     & $L_1$ Distance \\
                & threshold                  & 8*std          \\ \hline
Statistic model & is\_normalized             & False          \\
                & similarity measurement     & Inner Product  \\
                & threshold                  & 8*std          \\ \hline
Scann           & number of leaves           & 2000           \\
                & number of leaves to search & 100            \\
                & training sample size       & 250000         \\
                & reorder                    & 5000           \\ \hline
Active learning & $\alpha$      & 0.5            \\
                & $\beta$       & 0.5            \\
                & $\gamma$     & 1              \\
                & $\delta$      & 0.2            \\ \hline
LightGBM        & number of leaves           & 200            \\
                & objective                  & Binary         \\ \hline
\end{tabular}
\end{table}



\end{document}